\documentclass[twoside,11pt]{article}

\usepackage[preprint]{jmlr2e}
\usepackage{tikz}
\usepackage{todonotes}
\usepackage{amssymb}
\usepackage{pifont}
\usepackage{array}
\usepackage{bm}
\usepackage[shortlabels]{enumitem}

\usepackage{amsmath}
\usepackage{amssymb}
\usepackage{hyperref}

\newcommand{\cmark}{\ding{51}}%
\newcommand{\xmark}{}%

\jmlrheading{NA}{2023}{NA}{17/04}{NA}{ankur00a}{Ankur Ankan and Johannes Textor}

\ShortHeadings{pgmpy: A Python Toolkit for Bayesian Networks}{Ankan and Textor}
\firstpageno{1}

\begin{document}

\title{pgmpy: A Python Toolkit for Bayesian Networks}

\author{\name Ankur Ankan \email ankur.ankan@ru.nl
       \AND
       \name Johannes Textor \email johannes.textor@ru.nl \\
       \addr Institute of Computing and Information Sciences, Radboud University, Netherlands \\
       }

\editor{}

\maketitle

\begin{abstract}

	Bayesian Networks (BNs) are used in various fields for modeling,
	prediction, and decision making. pgmpy is a python package that
	provides a collection of algorithms and tools to work with BNs and
	related models. It implements algorithms for structure learning,
	parameter estimation, approximate and exact inference, causal
	inference, and simulations. These implementations focus on
	modularity and easy extensibility to allow users to quickly modify/add
	to existing algorithms, or to implement new algorithms for different use
	cases. pgmpy is released under the MIT License; the source code is
	available at: \url{https://github.com/pgmpy/pgmpy}, and the
	documentation at: \url{https://pgmpy.org}.
\end{abstract}

\begin{keywords}
  Bayesian Networks, Directed Acyclic Graphs, Causal Inference, Probabilistic Inference, Simulation, Structure Learning
\end{keywords}

\section{Introduction}
Bayesian Networks (BNs), also known as Belief Networks, and related models such
as Directed Acyclic Graphs (DAGs), Structural Equation Models (SEMs), and
Dynamic Bayesian Networks (DBNs) are used in a variety of applications
\citep{pourret2008} like healthcare \citep{kyrimi2021}, medicine
\citep{arora2019}, natural language processing \citep{goyal2008}, computational
biology \citep{needham2007,Chen2022}, robotics \citep{lazkano2007,
premebida2016, amiri2022}, neuroscience \citep{bielza2014}, and others. Some of
the most common tasks in these applications are learning model structure from
data (structure learning), estimating model parameters (parameter learning),
querying a model for conditional or marginal distributions (probabilistic
inference), causal effect estimation between variables (causal inference), and
simulating data under various generating processes. pgmpy provides algorithms
to perform all these tasks for discrete variable BNs, with a subset of
algorithms supporting continuous and mixed data. We describe the main features
and the implemented algorithms in detail in Section~\ref{sec:features}.

Numerous algorithms have been proposed for solving BN tasks and newer ones are
continuously being developed. Therefore, one of the main design goals of pgmpy is to
make it easy to modify or add to existing algorithms and allow users to
implement new algorithms quickly. To achieve this, pgmpy is written in pure
python for readability with a focus on code modularity. Most algorithm
implementations can accept custom user-defined functions and classes as
arguments. Each class of algorithms follows the class structure defined by its
abstract base class, which also defines helpful data structures. This allows
users to quickly implement new algorithms by inheriting these base classes.
Another benefit of this approach is that new algorithms can be plugged
back in and be used with any other feature. Apart from extensibility, another
priority has been reliability through high test coverage and a well-documented
code base.

\section{Main Features}
\label{sec:features}
\begin{figure}[t]
	\centering
	\begin{tikzpicture}[yscale=.86, inner sep=1pt]
	\tikzstyle{every node}=[align=left]
		\node (data)  at (0, 0) {Data};
		\node (dag)    at (5, 0) {DAG};
		\node (bn)    at (12, 0) {Parameterized BN};
		\node (test)  at (2.5, -0.8) {Model Testing};
		\node (iv)    at (3.7, -1.4) {Instrumental Variables};
		\node (as)    at (6, -0.8) {Adjustment Set};
		\node (sim)   at (9, -0.8) {Simulations};
		\node (ci)    at (9.8, -1.4) {Causal Inference};
		\node (pi)    at (11.2, -2) {Probabilistic Inference};
		\node (testp) at (14, -0.8) {Model Testing};
		\node (readwrite) at (12.8, -1.4) {Export};
		\node (read) at (12, 0.8) {Import};

		\draw[->] (data) edge [bend left=10] node [midway, above]{Structure Learning} (dag.west);
		\draw (dag.east) edge [->,bend left=10] node [midway, above]{Parameter Learning} (bn.west);
		\draw[->] (dag.south) -- (test.north);
		\draw[->] (dag.south) -- (iv.north);
		\draw[->] (dag.south) -- (as.north);
		\draw[->] (bn.south) -- (pi.north);
		\draw[->] (bn.south) -- (ci.north);
		\draw[->] (bn.south) -- (sim.north);
		\draw[->] (bn.south) -- (testp.north);
		\draw[->] (bn.south) -- (readwrite.north);
		\draw[->] (read.south) -- (bn.north);
	\end{tikzpicture}
	\caption{Possible Workflows in pgmpy for Bayesian Networks}
	\label{fig:workflow}
\end{figure}
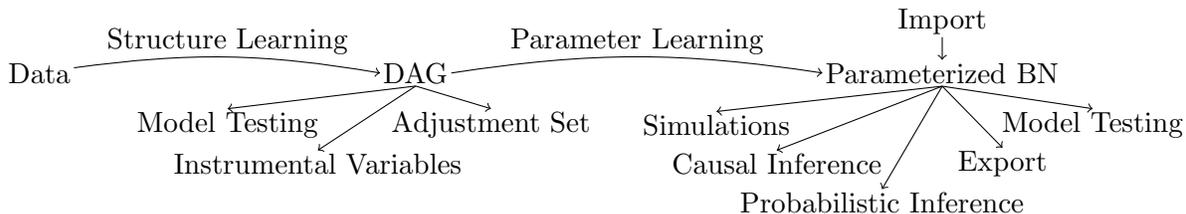
This section gives an overview of the main features and the implemented
algorithms. The algorithms only support discrete variables unless stated
otherwise.

\paragraph{Structure Learning (SL):}
Given a dataset, SL algorithms estimate the model structure. pgmpy implements
three general-purpose SL algorithms:
\begin{enumerate}[nosep]
	\item \textbf{PC:} A constraint-based algorithm that exploits the
		conditional independences (CIs) in the data to construct the
		model structure. Three variants of the algorithm are
		implemented: a) Original \citep{spirtes1993}, b) Stable
		\citep{colombo2014}, and c) Parallel \citep{le2019}. The
		following CI tests are available: a) chi-squared, b) G-test, c)
		Cressie-Read Power-Divergence, d) partial correlation, and e)
		residualization test \citep{ankan2022}. The partial correlation
		test can be used only for continuous data and the residualization
		test can be used for mixed data. Users can also specify custom
		tests as a function.
	\item \textbf{Hill-Climb Search (HC):} A score-based algorithm that
		iteratively makes local changes in the model structure to
		improve its score. The following scoring metrics are
		implemented: a) K2 Score \citep{cooper1992}, b) BDeu
		\citep{buntine1991}, c) BDs \citep{scutari2016}, d) BIC
		\citep{schwarz1978}, e) AIC \citep{akaike1974}. Users can also
		specify custom scoring methods.
	\item \textbf{Max-Min Hill-Climbing(MMHC):} The MMHC algorithm
		\citep{tsamardinos2006} combines constraint-based and
		score-based approaches by using the output of the MMPC algorithm
		\citep{tsamardinos2003}, which is a constraint-based algorithm, as
		the starting point for HC algorithm.

\end{enumerate}

Apart from the general purpose SL algorithms, pgmpy also implements the
 Chow-Liu algorithm \citep{chow1968} and 
 Tree Augmented Naive Bayes \citep{friedman1997} to learn tree structures.
Both algorithms are based on constructing a maximum spanning tree over the
variables using a (conditional) mutual information based metric as the edge
weights. Users can use mutual information (or its variants like adjusted or
normalized mutual information) as the edge weights or they can specify a custom edge
weight function.

\paragraph{Parameter Learning:}
The SL algorithms output a DAG, which can be used for causal inference analyses
like finding adjustment sets or instrumental variables
(Figure~\ref{fig:workflow}). However, for applications like inference and
simulations we also need to learn the model's parameters. The following
three methods are implemented to estimate the Conditional Probability
Distribution (CPD), also known as Conditional Probability Tables (CPTs), for the variables from (weighted) data:

\begin{enumerate}[nosep]
	\item \textbf{Maximum Likelihood (ML) Estimator:} Learns the ML estimates for the CPDs.
	\item \textbf{Bayesian Estimator:} Uses user-specified priors for each
		variable to do a Bayesian estimate of the CPDs. Shorthands are
		implemented to specify commonly used priors like Dirichlet,
		BDeu, or K2.
	\item \textbf{Expectation Maximization (EM):} Uses the EM algorithm to
		make Maximum Likelihood estimates in the presence of latent variables
		or missing data.
\end{enumerate}

\paragraph{Probabilistic Inference:}
Given a fully specified BN, the probabilistic inference algorithms allow users
to query the model for any conditional distribution, $ P(\bm{X} |
\bm{Y}=\bm{Y_i} )$. Currently, two algorithms are implemented for exact
inference, and a simulation-based approach is implemented for approximate
inference. All these algorithms allow users to specify combinations of hard
and virtual \citep{pearl1988} evidence.

\begin{enumerate}[nosep]
	\item \textbf{Variable Elimination (VE):} The
		VE algorithm \citep{zhang1994simple} conditions the model on the
		given evidence and iteratively sums out all the variables that
		are not in the query. The order in which the variables are
		summed out is crucial for the computational cost. pgmpy
		implements some algorithms and heuristics \citep{koller2009} to
		compute an efficient elimination order: 1) Optimized
		einsum-based \citep{smith2018} 2) Min Fill 3) Min Neighbours 4)
		Min Weight 5) Weighted Min Fill. Users can also specify a
		custom algorithm or a list of variable names.
	\item \textbf{Belief Propagation (BP):} The 
		BP algorithm \citep{judea1982} splits the query computation into two parts -- model
		calibration and query computation. It can be much faster than
		VE when multiple queries need to be made on the same model, as
		the calibration step does not need to be repeated.
	\item \textbf{Approximate Inference}: Depending on the model size
		and the query, both VE and BP can become intractable. In such
		cases, approximate inference can be used which computes the
		queried distribution by simulating data from the model.
\end{enumerate}

\paragraph{Causal Inference:}
The causal inference module provides features to estimate the causal effect between
a given exposure and an outcome variable. This can be done using:
\begin{enumerate}[nosep]
	\item \textbf{Instrumental Variables (IVs):} This method finds IVs and 
		conditional IVs \citep{van2015} that can be used with an
		IV-based estimator to get the estimates.
	\item \textbf{Adjustment Set:} This method finds the minimal adjustment set
		\citep{perkovic2018} that can be used with any
		statistical model to get the estimates.
\end{enumerate}
For continuous datasets, pgmpy implements a Two-Stage Least Squares (2SLS)
estimator for the IV method and a linear regression model for the adjustment
set method. For discrete datasets, estimates are computed using the
probabilistic inference engine. If users want to use a different statistical
model, they can use pgmpy to find the IVs and adjustment sets, and use any
statistical models package (e.g., statsmodels \citep{seabold2010}) for estimation.

\paragraph{Model Testing:} The model testing module provides methods to compare
the fit of different models to a given dataset. The following three metrics are
implemented: 1) Log-likelihood Score: Computes the log-likelihood of the
dataset given a fully-specified BN. 2) Structure Score: Computes the structure
score of a given model using one of the scoring metrics for the HC algorithm.
3) Correlation Score: Compares the implied correlations of the model with the
correlations present in the dataset and returns the F1-Score.

\paragraph{Simulations:}
The simulation method allows users to generate data from a fully specified BN
under any combination of the following conditions: a) Hard evidence, b) Virtual
evidence \citep{pearl1988}, c) Hard intervention, and d) Virtual intervention.

\paragraph{Exporting and Importing models:}
Finally, to interface with other packages pgmpy supports importing/exporting
models in the following file formats: a) Bayesian Interchange Format (BIF)
\citep{bif}, b) UAI \citep{uai}, c) XMLBIF \citep{xmlbif}, d) XMLBeliefNetwork
\citep{xbn}, and e) NET \citep{net}.

\begin{table}[t]
	\centering
	\begin{tabular}{|| c | c | c | c | c | c | c | c | c | c ||}
	\hline
		Package & SL & PL & EI & AI & Sim & CaI & cSim & MT & Ex/Im \\
	\hline \hline
		pgmpy & \cmark & \cmark & \cmark & \cmark & \cmark & \cmark & \cmark & \cmark & \cmark \\
	\hline
		pomegranate \citep{schreiber2017} & \cmark & \cmark & \xmark & \cmark & \cmark & \xmark & \xmark & \xmark & \xmark \\
	\hline
		pyBNesian \citep{atienza2022} & \cmark & \cmark & \xmark & \xmark & \cmark & \xmark & \xmark & \xmark & \xmark \\
	\hline
		bayespy \citep{luttinen2016} & \xmark & \cmark & \xmark & \cmark & \xmark & \xmark & \xmark & \xmark & \xmark \\
	\hline
		DoWhy \citep{dowhy_gcm} & \cmark & \xmark & \xmark & \xmark & \xmark & \cmark & \xmark & \cmark & \xmark \\
	\hline
		dagitty \citep{textor2016} & \xmark & \xmark & \xmark & \xmark & \cmark & \cmark & \xmark & \cmark & \cmark \\
	\hline 
		bnlearn \citep{scutari2010} & \cmark & \cmark & \cmark & \cmark & \cmark &\xmark & \xmark & \xmark & \cmark \\
	\hline 
		pcalg \citep{kalisch2012} & \cmark & \cmark & \cmark & \cmark & \cmark & \cmark & \xmark & \xmark & \cmark \\
	\hline
	\end{tabular}
	\caption{Feature comparison with other packages. We have excluded
		 causalnex \citep{causalnex_2021} and python bnlearn
		 \citep{bnlearn_2020} as they both extend pgmpy and hence have
		 all the functionality of pgmpy. SL=Structure Learning,
		 PL=Parameter Learning, EI=Exact Inference, AI=Approximate
		 Inference, Sim=Simulation, CaI=Causal Inference, cSim=Causal
		 Simulation, MT=Model Testing, Ex/Im=Export/Import}
	\label{table:compare}
\end{table}

\section{Conclusion and Future Work}
pgmpy provides a collection of tools and algorithms to work with BNs and
related models with a design focus on modularity and extensibility. Although
there are other python and R packages with similar features, most have
functionality for either probabilistic inference tasks or causal inference
tasks (Table~\ref{table:compare}). pgmpy combines these and allows
users to use methods from both these fields seamlessly.

In the future, we plan to add support for continuous variables to more
algorithms. We would also like to expand causal inference features by adding
new model classes (like Maximal Ancestral Graphs and Partial Ancestral Graphs)
along with causal discovery algorithms like Fast Causal Inference and Greedy
Equivalence Search.

\section*{Acknowledgements}
We would like to thank all the contributors of pgmpy. We would also like to
thank Google and Python Software Foundation for their support through the
Google Summer of Code program.

\bibliography{bibliography}

\end{document}